\documentclass[letterpaper, 10 pt, conference]{ieeeconf}  

\IEEEoverridecommandlockouts                              

\usepackage{graphics} 
\usepackage{epsfig} 
\usepackage{mathptmx} 
\usepackage{times} 
\usepackage{amsmath} 
\usepackage{amssymb}  
\usepackage{tabularx}
\usepackage{threeparttable}
\usepackage{booktabs}
\usepackage{multirow}
\makeatletter
\let\NAT@parse\undefined
\makeatother
\usepackage[
  colorlinks,
  linkcolor=magenta,
]{hyperref}

\title{\LARGE \bf
In-Rack Test Tube Pose Estimation Using RGB-D Data
}
\author{Hao Chen$^{1*}$, Weiwei Wan$^{1}$, Masaki Matsushita$^{2}$, Takeyuki Kotaka$^{2}$ and Kensuke Harada$^{13}$
\thanks{$^{1}$Department of System Innovation, Graduate School of Engineering Science, Osaka University, Toyonaka, Osaka, Japan.}
\thanks{$^{2}$H.U. Group Research Inst. G. K., Japan.}
\thanks{$^{3}$National Inst. of AIST, Japan.}
\thanks{$^{*}$Contact: Hao Chen, {\tt\small chen960216@gmail.com}}} 

\begin{document} 
\maketitle
\thispagestyle{empty}
\pagestyle{empty}

\begin{abstract}
Accurate robotic manipulation of test tubes in biology and medical industries is becoming increasingly important to address workforce shortages and improve worker safety. The detection and localization of test tubes are essential for the robots to successfully manipulate test tubes.  
In this paper, we present a framework to detect and estimate poses for the in-rack test tubes using color and depth data. The methodology involves the utilization of a YOLO object detector to effectively classify and localize both the test tubes and the tube racks within the provided image data. Subsequently, the pose of the tube rack is estimated through point cloud registration techniques. 
During the process of estimating the poses of the test tubes, we capitalize on constraints derived from the arrangement of rack slots. By employing an optimization-based algorithm, we effectively evaluate and refine the pose of the test tubes. This strategic approach ensures the robustness of pose estimation, even when confronted with noisy and incomplete point cloud data.


\end{abstract}

\section{Introduction}

Object detection and pose estimation play a vital role in various applications, like autonomous driving, industrial automation, and augmented reality. They function as the eyes for intelligent systems to identify objects and determine their position and orientation. This paper specially focuses on detecting and estimating poses of test tubes within a rack, which is a key task for automation in biology and medicine.

Detecting in-rack test tubes has unique challenges. The tubes in racks are often tightly spaced, resulting in partial occlusions from neighboring tubes that hinder detection and pose estimation. Furthermore, transparent or semi-transparent test tubes create complexity through refractive effects that degrade depth sensor data quality. Together, these intricacies have obstructed precise and robust test tube detection and localization. 

Many research has been conducted on object detection and pose estimation. 
Traditional methods first detect and estimate coarse poses for target objects using feature or template matching \cite{lowe1999object,hinterstoisser2011gradient}.
Then, the coarse pose is refined using iterative closest point (ICP) \cite{besl1992method}. These methods often require exhaustively searching the input data to match templates, which is inefficient and prone to failure with background clutter and sensor noise. Recent advance in machine learning used neural networks to extract features and can work well with the occlusion and clutter environment. Some end-to-end pose estimation networks like YOLO6d \cite{tekin2018real} can real-time estimate poses of objects in the clutter environment. However, learning-based methods require large amounts of training data to achieve good performance, which can be time-consuming and required considerable manual work to label data.

\begin{figure}[!tbp]
	\centering
	\includegraphics[width=\linewidth]{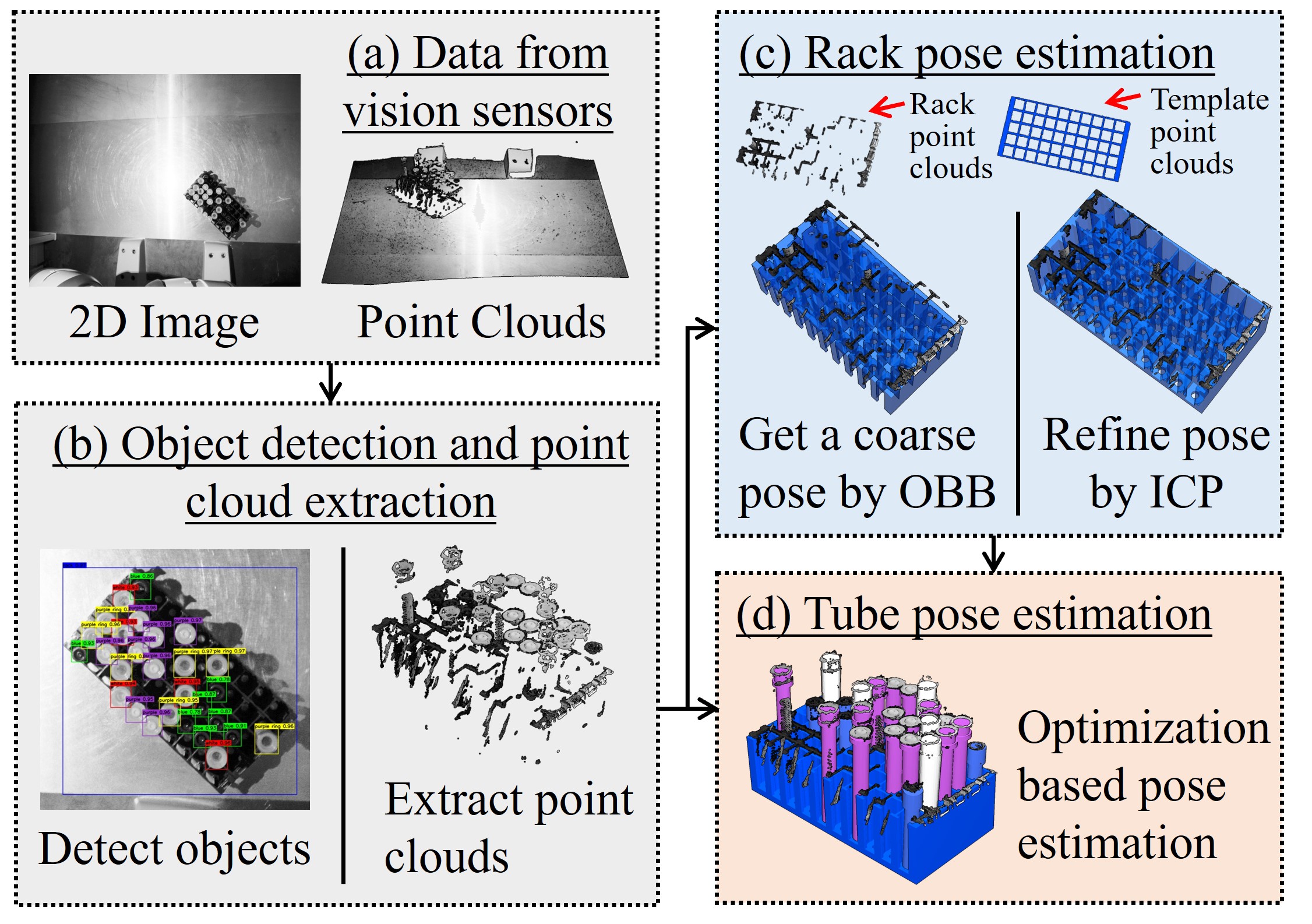}
 	\caption{The workflow of the system involves taking 2D images and point clouds as input. The outputs are the estimated poses of the test tubes and racks. A key innovation of our system is that we incorporate the correlation between each test tube and its corresponding rack slot as a prior constraint for test tube pose estimation. This enables robust and precise detection of test tubes even when the point cloud data of test tubes is noisy or incomplete.}
	\label{fig:flowchart}
\end{figure} 

In this paper, we developed a framework to detect and estimate poses of test tubes. 
The frameworks follows two-staged pipeline: a YOLO object detector first classify and localize test tubes and tube racks in color image. Then the pose of tube rack is analyzed by using traditional point cloud registration. After that, an optimization-based feature fitting method leverages extracted point clouds of test tubes and the pose of the test tube rack to estimate the pose of the test tubes. The method is efficient and precious. In some cases, point clouds of transparent/semi-transparent test tubes are corrupted, the proposed method can still estimate the poses for them. 


The contributions of this paper are two-fold. (1) We develop a framework to achieve effective in-rack test tube detection that can be utilized in various test tube manipulation tasks. (2) We propose an optimization-based feature fitting method that utilizes the pose of the test tube rack as a prior to estimate poses for test tubes, which enable accurate pose estimation for test tubes even in the presence of noise and incomplete point cloud data.

The remaining part of this paper is structured as follows: Section \ref{sec:related_work} reviews related work. Section \ref{sec:workflow} outlines the workflow of the proposed method. Section \ref{sec:technical_details} delivers technical details. Section \ref{sec:exp} shows experiments and analysis. Section \ref{sec:conclusion} draws conclusions.

\section{Related Work} 
\label{sec:related_work}
Most recent research treat object detection and pose estimation as two sequential tasks. Object detectors first classify and localize objects in an image, then pose estimators determine precise spatial locations and orientations for the detected objects. In this section, we review research on object detection and pose estimation. 

\subsection{Object Detection} Early object detection methods relied on handcrafted features \cite{dalal2005histograms}\cite{lowe2004distinctive} along with SVM classifiers \cite{malisiewicz2011ensemble}. However, traditional methods lacked of effective image representation, subsequently limiting accuracy. The advent of convolutional neural networks (CNNs) changed this landscape dramatically \cite{zou2023object}. R-CNN \cite{girshick2014rich} pioneered region proposals for localization coupled with CNNs for classification. Later one-stage detectors like YOLO \cite{redmon2016you} eliminated region proposals, using a single network for bounding box prediction and classification. Transformers like DETR have also emerged, replacing hand-designed anchors and non-maximum suppression (NMS) with attention mechanisms. Modern object detectors \cite{liu2021swin}\cite{wang2023yolov7} can achieve high accuracy while maintaining real-time performance.

In this paper, we employ YOLOv5 \cite{glenn_jocher_2020_4154370} to classify and localize the tube rack and various test tubes in 2D images. YOLOv5 excels in both speed and accuracy for object detection tasks. While learning-based object detectors require extensive training data to perform well, acquiring such data is a laborious and time-consuming process. In our previous work, we introduced an intuitive robotic data collection system that efficiently gathers training data for the object detector without requiring human intervention \cite{chen2023automatically}.

\subsection{Pose Estimation} 
Pose estimation methods can be broadly categorized into two approaches: feature-based methods and template-based methods.

In feature-based methods, early research focused on estimating poses by establishing correspondences between the 2D input image and the 3D object model through the matching of 2D handcrafted features \cite{lowe1999object, rothganger20063d}. However, these 2D feature matching methods tend to struggle when dealing with texture-less or symmetric objects. With the advent of consumer-grade depth sensors like Kinect and Realsense, feature-based methods such as \cite{rusu2008aligning, drost2010model} shifted towards establishing correspondences between point clouds and 3D models using 3D handcrafted features. Nonetheless, similar to methods based on 2D features, those relying on 3D features also exhibit performance limitations when dealing with objects exhibiting symmetrical properties.

In template-based methods, a prototype image representation—known as a template—is compared to various locations within the input image. At each location, a similarity score is computed, and the best match is determined by comparing these similarity scores. A popular example of this approach is LINEMOD \cite{hinterstoisser2011gradient}, where each template is represented by a map of color gradient orientations and surface normal orientations. However, it's worth noting that template matching methods are prone to issues of occlusion between objects, as the similarity score of the template is likely to be low if the object is occluded.

Recently, the integration of deep learning has significantly enhanced the performance of 6D pose estimation algorithms. PoseCNN \cite{xiang2017posecnn}, a deep learning architecture, garnered attention for its proficiency in predicting 3D object poses from single RGB images. Subsequently, methods like YOLO6d \cite{tekin2018real} have enabled real-time 6D pose estimation. Nonetheless, it's important to highlight that learning-based methods demand substantial training data to achieve satisfactory performance. This requirement can be time-consuming and often necessitates substantial manual effort for data labeling.

In this study, we proposed a feature-based method for in-rack test tube pose estimation.  
The method evaluates the test tube poses from point clouds based on a cylinder shaped feature fitting and novelly incorporating the correlation between each tube and its corresponding rack slot as a prior constraint. Even when the point cloud of a test tube is noisy or incomplete, our proposed method can still perform well. The experimental results clearly demonstrate the advantages of this method for detecting in-rack test tubes, as compared to prior work.




\section{Systematic Workflow}
\label{sec:workflow}
Fig.~\ref{fig:flowchart} illustrates the comprehensive workflow of the system. The system's inputs consist of two elements: a 2D image and point clouds. These data are sourced from a vision sensor. The resulting output comprises estimated poses for both the test tubes and the test tube rack.
The proposed vision system operates under three key assumptions: 1) The test tube rack is assumed to be opaque. 2) The test tubes are assumed to have their bottom centers approximately aligned with the central rack slot. This assumption stems from the fact that most tube racks incorporate a specific indentation at the bottom of each slot to facilitate test tube alignment.
3) The input gray images and point clouds have been previously calibrated by the manufacturer, ensuring that each pixel in the gray image corresponds to a point in the point cloud. If the image sensor and depth sensor lack calibration, the method outlined in \cite{basso2018robust} can be employed for calibration.

The system's initial step involves employing YOLO to detect both the test tubes and the test tube rack within the 2D image. YOLO is highly effective at identifying bounding boxes for these objects, subsequently enabling extraction of their corresponding point clouds. In our previous work \cite{chen2023automatically}, a data collection system was developed, combining robotic in-hand observation and data synthesis to automatically generate training data for the YOLO object detector. The object detector trained with dataset prepared by our method can achieve an impressive success rate of $98.8\%$ for the in-rack test tube detection task. 
Subsequent pre-processing steps, including outlier removal and the removal of any additional areas detected by YOLO, are applied to the extracted point clouds. Following this, the system estimates the pose of the tube rack by fitting a pre-collected point cloud template of the rack to the version obtained from the vision sensor. This fitting process involves utilizing the Oriented Bounding Box (OBB) and the Iterative Closest Point (ICP) algorithm. An initial transformation for the tube rack is determined by analyzing its OBB, and the ICP algorithm further refines the transformation to achieve higher accuracy.

Finally, the system evaluates the poses of the test tubes through an optimization-based approach that takes into account their point clouds and the pose of the tube rack. The method begins by projecting the point clouds of the test tubes onto the upper surface of the test tube rack, thereby identifying the corresponding slots for the tubes. Subsequently, the method determines the pose of each test tube by optimizing a rotation matrix. This matrix aligns the maximum number of points with the surface of a cylinder, which has a radius equivalent to that of the test tube.
The optimization-based pose estimation method employs the hole center's position and the test tube's radius to calculate the test tube's pose. Further details regarding this method are elaborated upon in the subsequent section.


\section{Optimization-based Pose Estimation Considering Rack Constraints}
\begin{figure}[!htbp]
	\centering
	\includegraphics[width=0.98\linewidth]{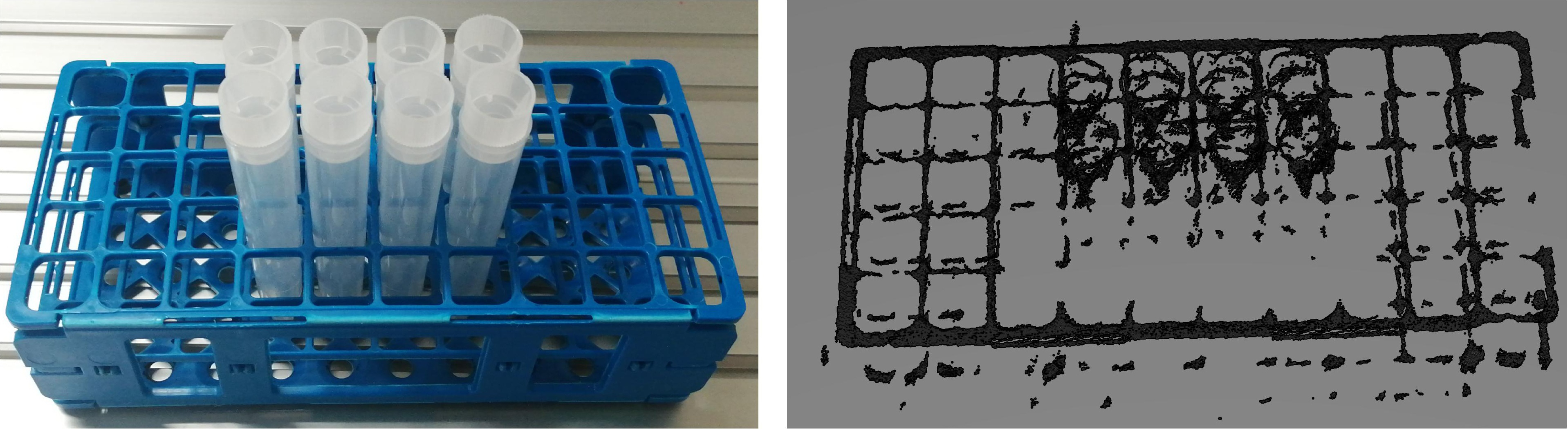}
 	\caption{The image on the left is an example of semi-transparent test tube detection. The right image shows the point cloud of the example.}
	\label{fig:example}
\end{figure}

In this section, the detail of the pose estimation is discussed. We use an example shown in Fig.~\ref{fig:example} to demonstrate the proposed method.

\subsection{Tube Rack Pose Estimation}
\begin{figure}[!htbp]
	\centering
	\includegraphics[width=\linewidth]{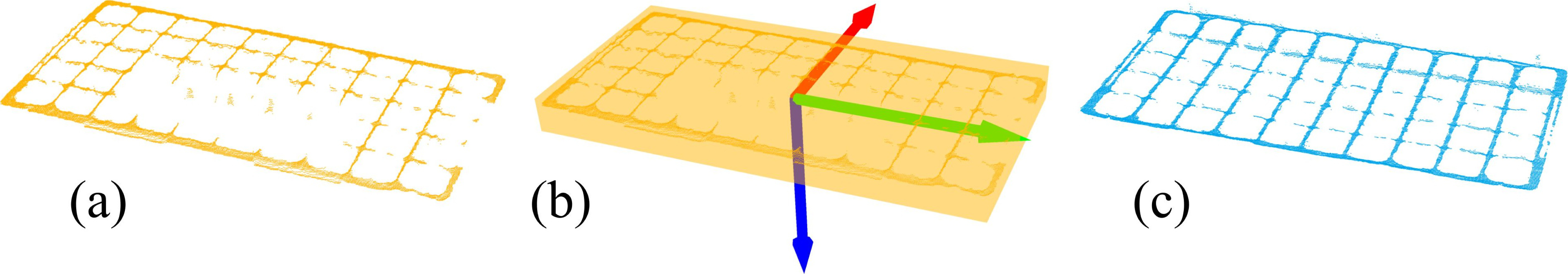}
 	\caption{(a) The extracted tube rack point cloud. (b) The OBB for the extracted tube rack point cloud. (c) The template tube rack point cloud used for point cloud registration.}
	\label{fig:boundingbox}
\end{figure}

In this subsection, we present our method for estimating the pose of the test tube rack using point cloud template matching. As shown in Fig.~\ref{fig:boundingbox} (c), we initially create a template using the top surface point cloud of the test tube rack. This template is chosen due to its dense point distribution and minimal occlusions. 

The process of estimating the test tube rack pose involves two key steps. 
The initial phase calculates an approximate transformation between the template point cloud and the point cloud acquired from the vision sensor. To achieve this, we employ the Oriented Bounding Box (OBB) algorithm, a well-established technique that identifies the box with the least area containing all the given points. For reference, Fig. \ref{fig:boundingbox} (a) and (b) provide an instance of the OBB applied to a point cloud cluster. The pose of the OBB serves as an initial estimate for the template point cloud's position.

\begin{figure}[!htbp]
	\centering
	\includegraphics[width=0.95\linewidth]{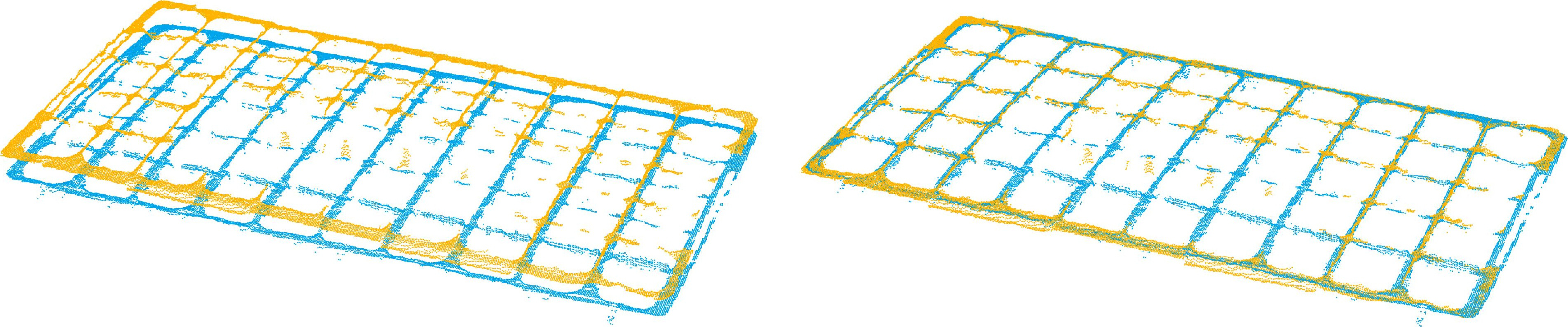}
 	\caption{The left image shows roughly estimated transformation by utilizing the OBB. The right image shows the pose refine by the ICP. The yellow point cloud is obtained from the vision sensor. The blue point cloud is the template.}
	\label{fig:icp}
\end{figure}

Subsequently, we enhance the transformation using the ICP algorithm. ICP is employed to iteratively minimize the reprojection error associated with fixed correspondences between the template point cloud and the captured point cloud until convergence is reached. This process ensures an accurate estimation of the test tube rack's pose. To visualize the results, Fig. ~\ref{fig:icp} demonstrates the rough estimate derived from OBB and the refined estimate achieved through ICP. Notably, the yellow point cloud represents data from the vision sensor, while the blue point cloud depicts the template. Once the pose of the test tube rack is determined, we can subsequently deduce the poses of the individual test tubes from this information. 

\subsection{Optimization-Based Tube Pose Estimation}
\label{sec:technical_details}
\begin{figure}[!htbp]
	\centering
	\includegraphics[width=0.95\linewidth]{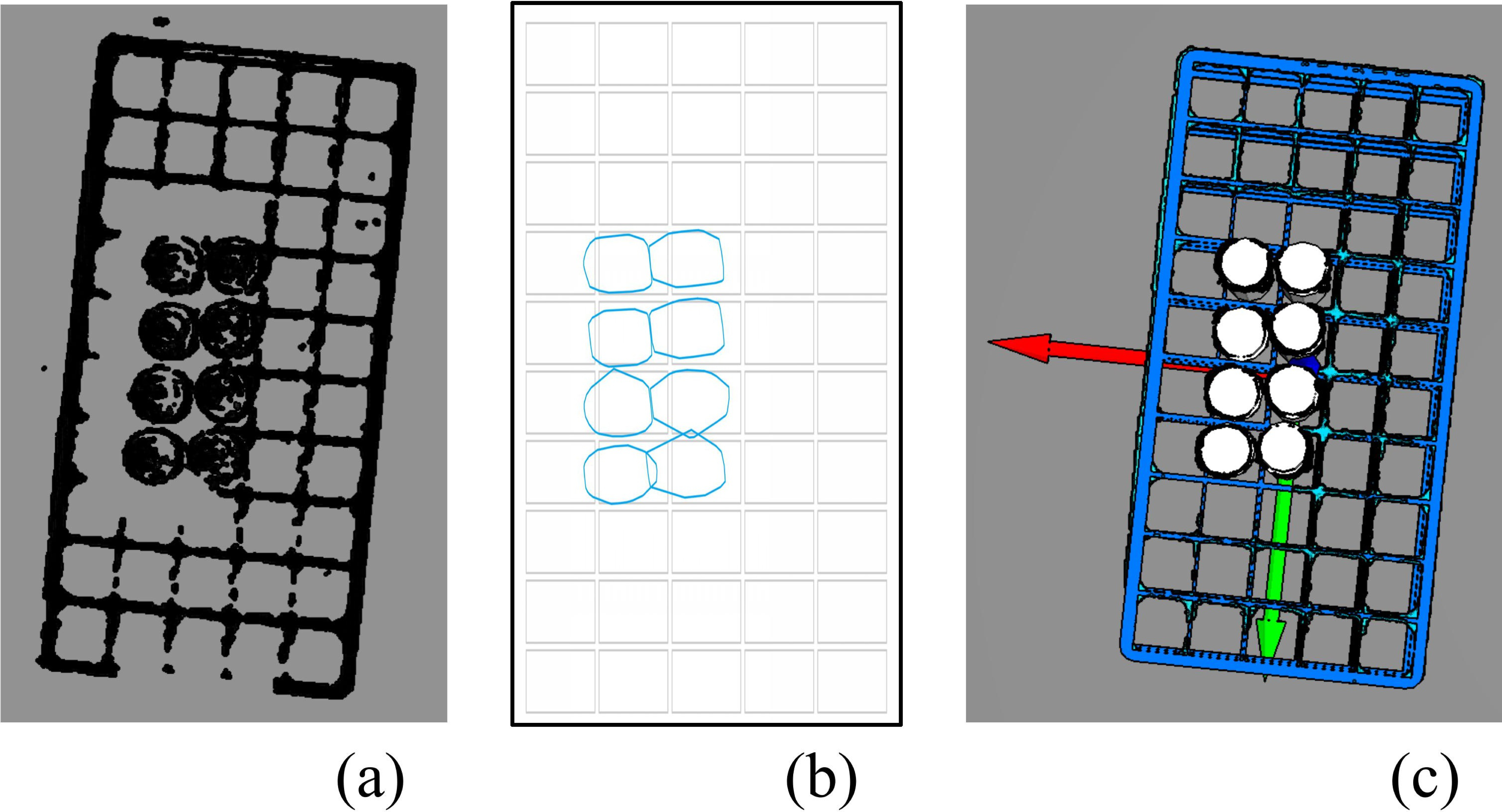}
 	\caption{(a) The point cloud of the example viewed from the top view. (b) The projection of convex hull of point cloud onto the top surface of the test tube rack. (3) The result using the proposed pose estimation method.}
	\label{fig:final}
\end{figure}

In this subsection, we present the method to estimate the poses of test tubes. The method firstly finds out the containing hole for each test tube by projecting the convex hull of its point cloud on the top surface of the test tube rack. Since we have already known the pose of the test tube rack, the position of each hole on the rack can also be concluded from it. Fig.~\ref{fig:final}(b) is the projection result of the example. The gray rectangles are the hole. The blue convex shapes are the projection of the convex hull of the test tube point cloud. The test tube is located in the hole containing the largest partition of the area of the its convex shape.

Support we have $N$ test tubes on the rack. The point cloud of $i$th test tube is notated as $\mathbf{P}^{i}=\{\mathbf{p}_1^{i}, \ldots, \mathbf{p}_j^{i}, \ldots \mathbf{p}_{M_i}^i\}$,  where 
$\mathbf{p}_j^{i} \in \mathbb{R}^{3}$, 
$M_i$ is the number of points in the $i$th point cloud. 
The radius of $i$th test tube is $r_i$.
Suppose the dimension of holes of the test tube rack are the same, and the notions of length, width, height are $2L$, $2W$, $H$ respectively.

\begin{figure}[!htbp]
	\centering
	\includegraphics[width=0.75\linewidth]{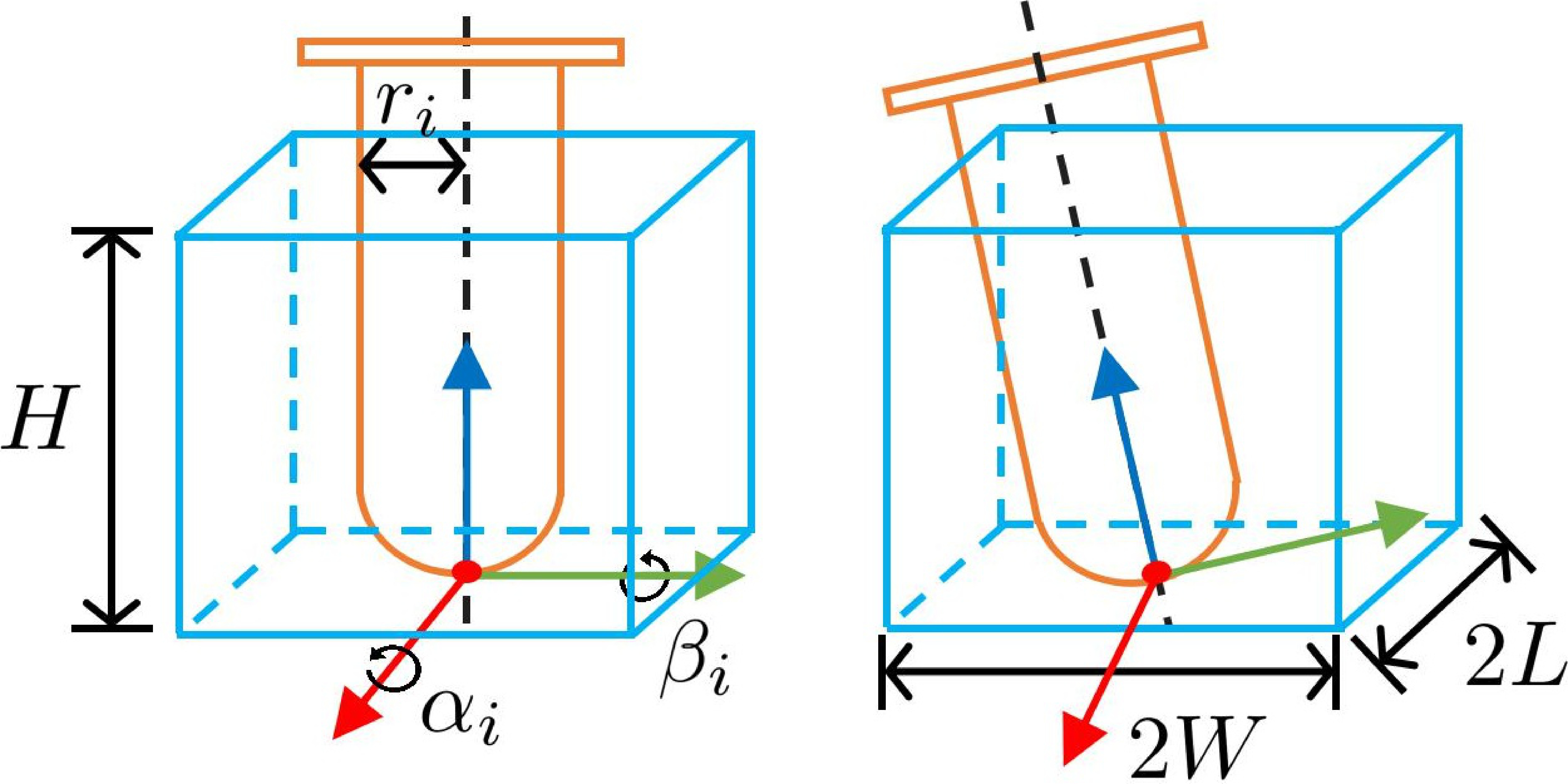}
 	\caption{The demonstration of the test tube in the rack hole. The red, green, blue arrows represent the x, y, z-axis of the test tube local coordinate. The left image shows an ideal case: the central axis of the transparent test tube and the central axis of the hole containing the test tube are coincident. The right image shows a case that the test tube reaches the boundary of the test tube hole.}
	\label{fig:geom}
\end{figure}

As shown in Fig.~\ref{fig:geom}, the local coordinate of the test tube is attached to the central bottom. 
To estimate the pose of the $i$th test tube, we use two parameters $\alpha_i$, $\beta_i$ to represent the orientation, where $\alpha_i$, $\beta_i$ are the Euler angles of $x$ axis and $y$ axis of the $i$th test tube coordinate. Then the orientation of the $i$th test tube can be defined as:
\begin{equation}
    \begin{aligned}
        \mathbf{R}(\alpha_i,\beta_i) &=
        \begin{bmatrix}
    cos\beta_i & 0 & sin\beta_i \\
    0 & 1 & 0 \\
    -sin\beta_i & 0 & cos\beta_i
    \end{bmatrix}
        \begin{bmatrix}
    1 & 0 & 0 \\
    0 & cos\alpha_i & -sin\alpha_i \\
    0 & sin\alpha_i & cos\alpha_i
    \end{bmatrix}
     \\
    &= \begin{bmatrix}
    cos\beta_i & sin \beta_i sin\alpha_i & sin\beta_i cos\alpha_i \\
    0 & cos\alpha_i & -sin\alpha_i \\
    -sin\beta_i & \cos{\beta_i} sin\alpha_i & cos\beta_i cos\alpha_i
    \end{bmatrix}
    \end{aligned} 
\end{equation}

\begin{figure}[!htbp]
	\centering
	\includegraphics[width=0.9\linewidth]{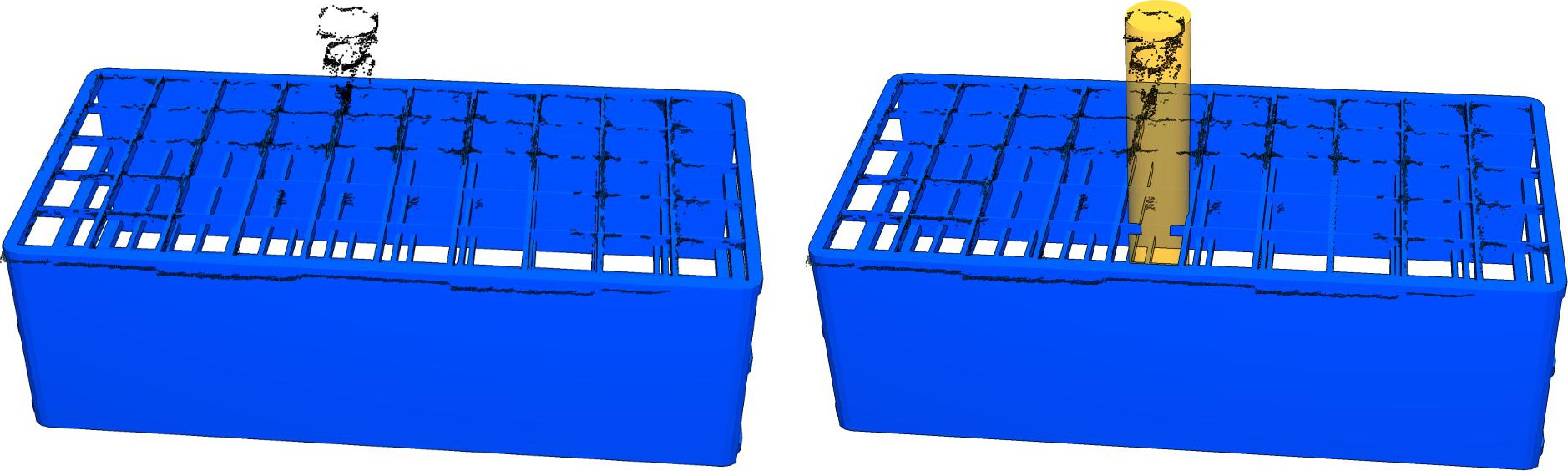}
 	\caption{The left image shows an incomplete point cloud of the transparent test tube. The right image shows the fitted cylinder for the point cloud.}
	\label{fig:rack}
\end{figure}

The distance between the $\mathbf{p}_j$ to the line segment that starts from origin point and points to $z$ direction of $i$th test tube coordinate is represented as:
\begin{equation}
    D(\alpha_i,\beta_i, \mathbf{p}_j,\mathbf{o}_i) =
    || (\mathbf{o}_i-\mathbf{p}_j)\times \begin{bmatrix}
        sin\beta_i cos\alpha_i \\-sin\alpha_i  \\ cos\beta_i cos\alpha_i
    \end{bmatrix}|| 
\end{equation}
where  $\mathbf{o}_i$ is the origin point of the $i$th test tube coordinate in the world coordinate, which can be concluded from the pose of the test tube rack. We use a cylinder that has the same radius as the test tube to represent the simplified geometry feature for the test tube. The pose of the cylinder is considered to be the same as the test tube.  The orientation of the cylinder is controlled by the parameter $\alpha$ and $\beta$. Although some parts of the point cloud of the transparent test tube miss, the remaining part of the point cloud still has enough features for the cylinder to fit. 
Fig.~\ref{fig:rack} shows an example of using the cylinder to fit the point cloud.
We suppose that the cylinder is in the correct pose when the point cloud is located near the surface of the cylinder.  By rotating the cylinder, we can determine the optimal pose for the cylinder that point cloud is well distributed near its surface. The following equation is used to find out the pose.
\begin{equation}
    \begin{aligned}
\min_{\alpha,\beta \in (-\pi,\pi]} \quad &  \sum_{k=1}^{M_i}(D(\alpha_i,\beta_i, \mathbf{p}_k, \mathbf{o}_i) - r_i)
    \end{aligned}
\end{equation} 
We define $\alpha_i^*$ and $\beta_i^*$ as the solution to minimize the distance between extracted point cloud and the surface of the simplified cylinder model. The final pose of $i$th test tube is 
\begin{equation}
    i\text{th test tube pose} = \begin{bmatrix}
        \mathbf{R}(\alpha_i^*,\beta_i^*) & \mathbf{o}_i \\
        \mathbf{0} & 1
    \end{bmatrix}
\end{equation}
Fig.~\ref{fig:final}(c) shows a result of the pose estimation of the example.
In addition, we need to define a function to reject wrong cases. For the wrong cases that point clouds of test tubes are seriously corrupted, the simplified cylinder geometry feature no longer exists in the point cloud. So the estimated pose is meaningless. The meaningless estimated pose can be effectively found by examining the geometry constraints between the rack hole and test tube. The geometry constraints tell us the $\alpha_i$ and $\beta_i$ should be bound in the following range so that the $i$th test tube and test tube rack have no overlap.

\begin{equation}
\begin{aligned}
    &tan(\alpha_i) (W-\frac{r_i}{sin\alpha_i}) > H\\
    &tan(\pi-\alpha_i) (W-\frac{r_i}{sin(\pi-\alpha_i)}) > H\\
    &tan(\beta_i) (L-\frac{r_i}{sin\beta_i}) > H\\
    &tan(\pi-\beta_i) (L-\frac{r_i}{sin(\pi-\beta_i)}) > H 
\end{aligned}
\end{equation}

\section{Experiments}
\label{sec:exp}
\subsection{Setup and Evaluation Metrics}
\begin{figure}[!htbp]
	\centering
	\includegraphics[width=\linewidth]{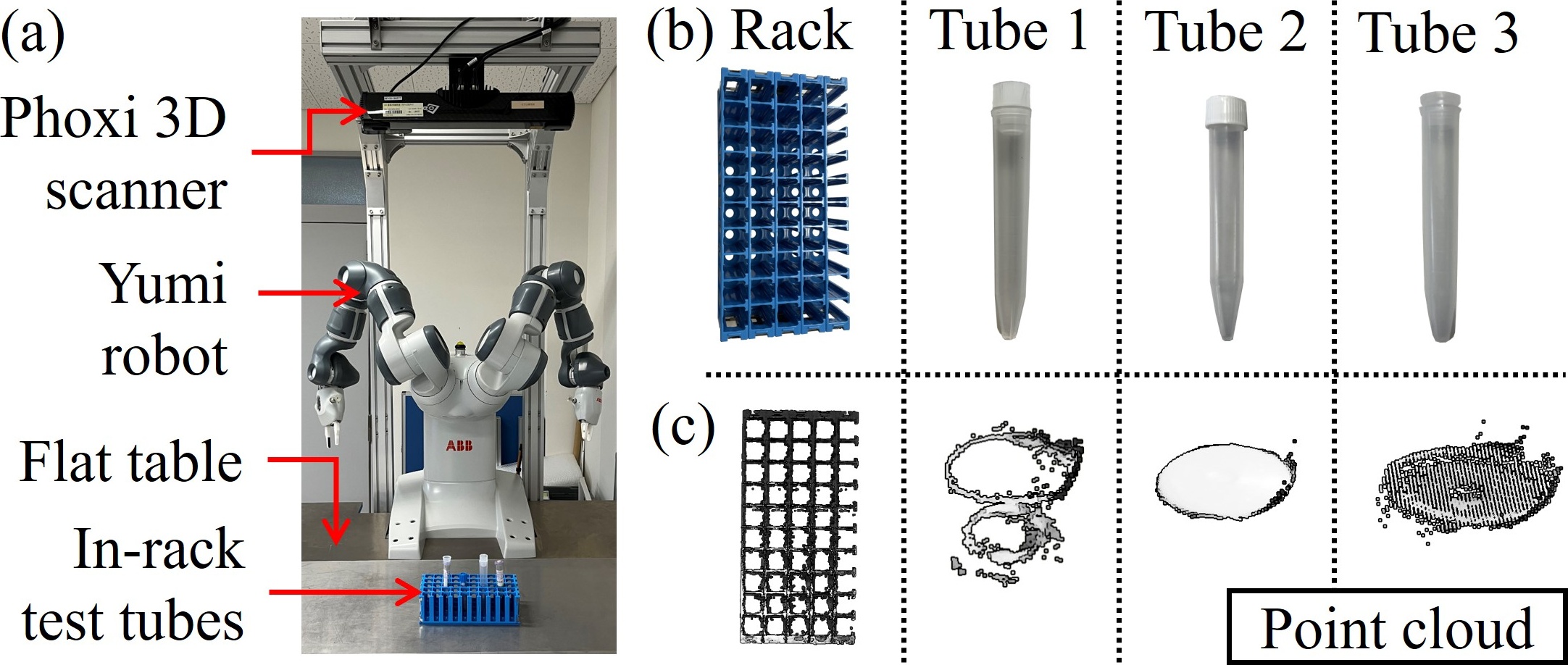}
 	\caption{(a) The system configuration. (b) The image of test tubes and
rack. (c) The point cloud of test tubes and rack obtained with the Phoxi 3D scanner above.}
	\label{fig:exp}
\end{figure}

In this section, we have assessed our proposed pose estimation algorithm for test tubes. Fig. \ref{fig:exp}(a) depicts the experimental setup employed for evaluating the accuracy of our pose estimation approach. In this configuration, a Photoneo Phoxi M 3D scanner is positioned above. The test tubes and rack is positioned on a flat surface table. 

We conducted evaluations on three types of test tubes. These test tubes, along with their corresponding point clouds, are illustrated in Fig. \ref{fig:exp}(b). Among these, the ``Tube 1'' posed the greatest challenge due to its semi-transparent materials, resulting in corrupted and incomplete point cloud data. In contrast, the remaining two test tubes are with opaque tube caps, resulting in a relatively simpler detection process. 

\begin{figure}[!htbp]
	\centering
	\includegraphics[width=\linewidth]{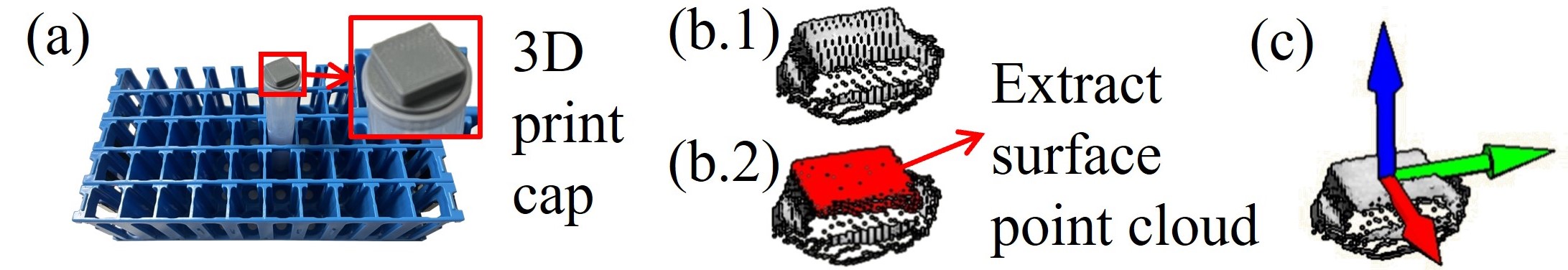}
 	\caption{(a) The custom 3D printed caps. (b.1) The point clouds of a 3d printed cap. (b.2) The point clouds in red are extracted using plane segmentation algorithm from Open3D library \cite{zhou2018open3d}. (c) A ground truth pose generated by point cloud registration. The red, green and blue arrows represent the X, Y, Z axes, respectively. }
	\label{fig:groundtruth}
\end{figure}

To evaluate the poses of the test tubes, we designed custom 3D-printed caps, as shown in Fig. \ref{fig:groundtruth}(a). Each cap featured a square surface at its top. The square top surface can be easily extracted by employing plane segmentation algorithms. 
After extraction, these square features were employed as target point clouds, which served as a basis for fitting a square point cloud template via the ICP algorithm. This process yielded accurate transformations that were employed as the ground truth for the orientations of the test tubes.
To further determine the translation of the test tubes, we measured the physical distance between the center of the square top surface and the origin point of the test tubes. 
With the ground truth data in hand, we proceeded to replace the 3D-printed caps with the original caps of the test tubes, while maintaining the test tube's pose unchanged. Subsequently, we performed scans to capture point cloud data for the test tubes with their original tube caps. This approach enabled a comprehensive comparison of our pose estimation algorithm's performance under realistic conditions. Note that while a highly accurate sensor isn't strictly required for our pose estimation algorithm, for the purpose of our evaluations, we opted for a high-precision sensor to obtain a point cloud in order to acquire a point cloud with minimal distortion. This choice ensured the establishment of a precise ground truth for the poses of the test tubes.
  
Regarding the evaluation metrics, we assessed the average rotational and translational errors along each axis. To compute the rotational error, we measured the angle between the quaternion of the attitude parameter (converted to Euler angles) and the correct angle for each rotation axis. Notably, since we defined the rotational symmetric axis for the test tube as the Z-axis, our assessment solely focused on the rotational error along the X and Y axes. On the other hand, the translation error was determined as the disparity between the translations observed along each axis.

\subsection{Comparison with ICP}
The results of the pose estimation for the test tubes are presented in Table \ref{tab:result_comp}. To facilitate a comprehensive evaluation, we have chosen the popular pose estimation technique of point cloud registration for comparison. This technique involves utilizing the global registration + ICP algorithm from the Open3D library, which efficiently establishes transformation between the template point cloud and the target point cloud. The template point cloud is generated in accordance with the CAD model of the respective test tubes.

When considering the pose estimation results for "Tube 1," it is evident that our proposed method significantly outperforms the global registration + ICP approach. This notable improvement can be attributed to the substantial dissimilarity between the realistic point cloud of the semi-transparent "Tube 1" and the point cloud template derived from the CAD model. This disparity often leads to the failure of global registration, consequently yielding incorrect outcomes. On the contrary, for "Tube 2" and "Tube 3," the presence of opaque caps in their point clouds allows for facile identification through the cap features. Moreover, when evaluating the rotational error, incorporating constraints derived from the rack slot further reduces the rotational error. Our strategic consideration of rack-based constraints aids in mitigating the impact of noise.

However, when examining the translational error associated with "Tube 3," our method demonstrates a relatively larger discrepancy in translational accuracy compared to the ICP method. This variance can be attributed to our reliance on the assumption that the bottom centers of the test tubes are roughly aligned with the central rack slot. In future iterations, we plan to refine our algorithm by removing this constraint, thereby enhancing its versatility and applicability across various scenarios.



\begin{table}[!htbp]
\caption{Evaluation results on rotational and translational errors}
\label{tab:result_comp}
\centering

\begin{threeparttable}
\begin{tabular}{ll|cc|ccc}
\toprule
&  & \multicolumn{2}{c|}{R.E.} &\multicolumn{3}{c}{Translational Error}\\
\cmidrule{3-7}
&  Method & $R_x$ &  $R_y$ & $T_x$ &  $T_y$  &  $T_z$ \\
\midrule
\multirow{2}{*}{Tube 1}
& Our  & $1.26^{\circ}$  &  $1.15^{\circ}$  &  $1.18$mm & $2.80$mm & $0.35$mm\\
& ICP & $43.4^{\circ}$  &  $28.2^{\circ}$  &  $21.1$mm   & $56.7$mm & $49.7$mm\\  
\midrule
\multirow{2}{*}{Tube 2}
& Our  & $ 0.97^{\circ}$  &  $ 0.88^{\circ}$  &  $0.69$mm   & $2.20$mm & $0.37$mm\\
& ICP & $2.24^{\circ}$  &  $1.77^{\circ}$  &  $1.38$mm   & $4.44$mm & $0.25$mm\\  
\midrule
\multirow{2}{*}{Tube 3}
& Our  & $1.06^{\circ}$  &  $ 0.93^{\circ}$  &  $0.88$mm   & $3.26$mm & $0.74$mm\\
& ICP & $1.06^{\circ}$  &  $0.95^{\circ}$  &  $1.44$mm   & $1.31$mm & $0.75$mm\\  
\bottomrule
\end{tabular}
  \begin{tablenotes}
  \item[Note 1] The rotational and translational errors are the mean of 20 trial for each test tube.
  \item[Note 2] The method ``ICP'' refers to global registration + ICP 
  \item[Note 3] R. E. = Rotational Error.
  \end{tablenotes}
\end{threeparttable}
\end{table}


\subsection{Time Costs}
In this subsection, we evaluate the time costs of our proposed method, including the time for the object detection in 2D image and the subsequent pose estimation on 3D point cloud. The specifications of our computing setup are as follows: an Intel i9-13900K CPU and an Nvidia RTX 4090 GPU. Note that the speed of the object detection stage primarily hinges on the GPU device, while the pose estimation stage is influenced by the CPU device. The results are shown in Table \ref{tab:time_con}. 
It includes the time taken for object detection per image,, rack pose estimation time per rack, the tube estimation time per tube using our method, and tube estimation time per tube through point cloud registration, utilized for comparison.

The results illustrate that the most time-consumption part in our framework is the rack pose estimation component, which necessitates performing the ICP algorithm on an extensive set of point clouds to ensure the precision of pose estimation. Another significant observation is the notable speed enhancement achieved by our proposed tube pose estimation method, showcasing an approximate threefold increase in speed compared to the traditional point cloud registration approach. This distinction becomes particularly significant when dealing with a substantial number of test tubes placed on the rack, a scenario frequently encountered in real-world conditions. 

\begin{table}[!htbp]
\caption{Time costs}
\label{tab:time_con}
\centering

\begin{threeparttable}
\begin{tabular}{l|c}
\toprule
Component &  Average Time per Instance\\
\midrule
Object Detection & $0.096$s \\
Rack Pose Estimation & $0.191$s  \\
Tube Pose Estimation (Our) & $0.012$s  \\
Tube Pose Estimation (ICP) & $0.036$s  \\ 
\bottomrule
\end{tabular}
\begin{tablenotes}
  \item[Note 1] The dimension of the input images for the object detection is $1376\times1376$.
  \item[Note 2] The average time is the mean of 10 in-rack test tube detection trials.
  \item[Note 3] The average time includes the time for point cloud outlier removing and downsampling.
  \end{tablenotes}
\end{threeparttable}

\end{table}

\section{CONCLUSIONS}
\label{sec:conclusion}
In this paper, we developed the framework for the detection and pose estimation of in-rack test tubes.  
Through a two-staged process, we are able to achieve good accuracy and efficiency in detecting both the rack structure and test tubes. The experimental result showed that our proposed method exhibits substantial improvements in both accuracy and computational efficiency by comparing with traditional point cloud registration method. These results underscore the significant potential of our approach in advancing the field of in-rack test tube detection.

One potential limitation of the framework is that inaccuracies in rack pose estimation could consequently impact the accuracy of tube pose estimation. For future work, the framework could be refined by focusing solely on analyzing the rack slots from the point cloud data and using the estimated rack slot information as priors for tube pose estimation. This approach would eliminate the need for accurate rack pose estimation.

\addtolength{\textheight}{-12cm}   







\bibliographystyle{IEEEtran}
\bibliography{root}

\begin{thebibliography}{10}
\providecommand{\url}[1]{#1}
\csname url@samestyle\endcsname
\providecommand{\newblock}{\relax}
\providecommand{\bibinfo}[2]{#2}
\providecommand{\BIBentrySTDinterwordspacing}{\spaceskip=0pt\relax}
\providecommand{\BIBentryALTinterwordstretchfactor}{4}
\providecommand{\BIBentryALTinterwordspacing}{\spaceskip=\fontdimen2\font plus
\BIBentryALTinterwordstretchfactor\fontdimen3\font minus
  \fontdimen4\font\relax}
\providecommand{\BIBforeignlanguage}[2]{{%
\expandafter\ifx\csname l@#1\endcsname\relax
\typeout{** WARNING: IEEEtran.bst: No hyphenation pattern has been}%
\typeout{** loaded for the language `#1'. Using the pattern for}%
\typeout{** the default language instead.}%
\else
\language=\csname l@#1\endcsname
\fi
#2}}
\providecommand{\BIBdecl}{\relax}
\BIBdecl

\bibitem{lowe1999object}
D.~G. Lowe, ``Object recognition from local scale-invariant features,'' in
  \emph{Proceedings of the seventh IEEE international conference on computer
  vision}, vol.~2.\hskip 1em plus 0.5em minus 0.4em\relax Ieee, 1999, pp.
  1150--1157.

\bibitem{hinterstoisser2011gradient}
S.~Hinterstoisser, C.~Cagniart, S.~Ilic, P.~Sturm, N.~Navab, P.~Fua, and
  V.~Lepetit, ``Gradient response maps for real-time detection of textureless
  objects,'' \emph{IEEE transactions on pattern analysis and machine
  intelligence}, vol.~34, no.~5, pp. 876--888, 2011.

\bibitem{besl1992method}
P.~J. Besl and N.~D. McKay, ``Method for registration of 3-d shapes,'' in
  \emph{Sensor fusion IV: control paradigms and data structures}, vol.
  1611.\hskip 1em plus 0.5em minus 0.4em\relax Spie, 1992, pp. 586--606.

\bibitem{tekin2018real}
B.~Tekin, S.~N. Sinha, and P.~Fua, ``Real-time seamless single shot 6d object
  pose prediction,'' in \emph{Proceedings of the IEEE conference on computer
  vision and pattern recognition}, 2018, pp. 292--301.

\bibitem{dalal2005histograms}
N.~Dalal and B.~Triggs, ``Histograms of oriented gradients for human
  detection,'' in \emph{2005 IEEE computer society conference on computer
  vision and pattern recognition (CVPR'05)}, vol.~1.\hskip 1em plus 0.5em minus
  0.4em\relax Ieee, 2005, pp. 886--893.

\bibitem{lowe2004distinctive}
D.~G. Lowe, ``Distinctive image features from scale-invariant keypoints,''
  \emph{International journal of computer vision}, vol.~60, pp. 91--110, 2004.

\bibitem{malisiewicz2011ensemble}
T.~Malisiewicz, A.~Gupta, and A.~A. Efros, ``Ensemble of exemplar-svms for
  object detection and beyond,'' in \emph{2011 International conference on
  computer vision}.\hskip 1em plus 0.5em minus 0.4em\relax IEEE, 2011, pp.
  89--96.

\bibitem{zou2023object}
Z.~Zou, K.~Chen, Z.~Shi, Y.~Guo, and J.~Ye, ``Object detection in 20 years: A
  survey,'' \emph{Proceedings of the IEEE}, 2023.

\bibitem{girshick2014rich}
R.~Girshick, J.~Donahue, T.~Darrell, and J.~Malik, ``Rich feature hierarchies
  for accurate object detection and semantic segmentation,'' in
  \emph{Proceedings of the IEEE conference on computer vision and pattern
  recognition}, 2014, pp. 580--587.

\bibitem{redmon2016you}
J.~Redmon, S.~Divvala, R.~Girshick, and A.~Farhadi, ``You only look once:
  Unified, real-time object detection,'' in \emph{Proceedings of the IEEE
  conference on computer vision and pattern recognition}, 2016, pp. 779--788.

\bibitem{liu2021swin}
Z.~Liu, Y.~Lin, Y.~Cao, H.~Hu, Y.~Wei, Z.~Zhang, S.~Lin, and B.~Guo, ``Swin
  transformer: Hierarchical vision transformer using shifted windows,'' in
  \emph{Proceedings of the IEEE/CVF international conference on computer
  vision}, 2021, pp. 10\,012--10\,022.

\bibitem{wang2023yolov7}
C.-Y. Wang, A.~Bochkovskiy, and H.-Y.~M. Liao, ``Yolov7: Trainable
  bag-of-freebies sets new state-of-the-art for real-time object detectors,''
  in \emph{Proceedings of the IEEE/CVF Conference on Computer Vision and
  Pattern Recognition}, 2023, pp. 7464--7475.

\bibitem{glenn_jocher_2020_4154370}
\BIBentryALTinterwordspacing
G.~Jocher \emph{et~al.}, ``{ultralytics/yolov5: v3.1 - Bug Fixes and
  Performance Improvements},'' Oct. 2020. [Online]. Available:
  \url{https://doi.org/10.5281/zenodo.4154370}
\BIBentrySTDinterwordspacing

\bibitem{chen2023automatically}
H.~Chen, W.~Wan, M.~Matsushita, T.~Kotaka, and K.~Harada, ``Automatically
  prepare training data for yolo using robotic in-hand observation and
  synthesis,'' \emph{arXiv preprint arXiv:2301.01441}, 2023.

\bibitem{rothganger20063d}
F.~Rothganger, S.~Lazebnik, C.~Schmid, and J.~Ponce, ``3d object modeling and
  recognition using local affine-invariant image descriptors and multi-view
  spatial constraints,'' \emph{International journal of computer vision},
  vol.~66, pp. 231--259, 2006.

\bibitem{rusu2008aligning}
R.~B. Rusu, N.~Blodow, Z.~C. Marton, and M.~Beetz, ``Aligning point cloud views
  using persistent feature histograms,'' in \emph{2008 IEEE/RSJ international
  conference on intelligent robots and systems}.\hskip 1em plus 0.5em minus
  0.4em\relax IEEE, 2008, pp. 3384--3391.

\bibitem{drost2010model}
B.~Drost, M.~Ulrich, N.~Navab, and S.~Ilic, ``Model globally, match locally:
  Efficient and robust 3d object recognition,'' in \emph{2010 IEEE computer
  society conference on computer vision and pattern recognition}.\hskip 1em
  plus 0.5em minus 0.4em\relax Ieee, 2010, pp. 998--1005.

\bibitem{xiang2017posecnn}
Y.~Xiang, T.~Schmidt, V.~Narayanan, and D.~Fox, ``Posecnn: A convolutional
  neural network for 6d object pose estimation in cluttered scenes,''
  \emph{arXiv preprint arXiv:1711.00199}, 2017.

\bibitem{basso2018robust}
F.~Basso, E.~Menegatti, and A.~Pretto, ``Robust intrinsic and extrinsic
  calibration of rgb-d cameras,'' \emph{IEEE Transactions on Robotics},
  vol.~34, no.~5, pp. 1315--1332, 2018.

\bibitem{zhou2018open3d}
Q.-Y. Zhou, J.~Park, and V.~Koltun, ``Open3d: A modern library for 3d data
  processing,'' \emph{arXiv preprint arXiv:1801.09847}, 2018.

\end{thebibliography}

\end{document}